# TPPoet: Transformer-Based Persian Poem Generation using Minimal Data and Advanced Decoding Techniques

*Amir Panahandeh[1]\**, *Hanie Asemi[2], Esmaeil Nourani[2]*

[1]*Faculty of Electrical and Computer Engineering, University of Tabriz, Iran*
[2]*Faculty of Information Technology and Computer Engineering, Azarbaijan Shahid Madani University, Tabriz, Iran[2]*

## Abstract

Recent advances in language models (LMs), have demonstrated significant efficacy in tasks related to the arts and humanities. While LMs have exhibited exceptional performance across a wide range of natural language processing tasks, there are notable challenges associated with their utilization on small datasets and their ability to replicate more creative human capacities. In this study, we aim to address these challenges by training a Persian classical poetry generation model using a transformer architecture on a specialized dataset with no pretraining. Additionally, we propose a novel decoding method to enhance coherence and meaningfulness in the generated poetry, effectively managing the tradeoff between diversity and quality. Furthermore, the results of our training approach and the proposed decoding method are evaluated through comprehensive set of automatic and human evaluations and showed its superior capability to generate coherent and meaningful poetry in compare to other decoding methods and an existing Persian large language model (LLM).

***Keywords:*** *Transformers, Generative Art, Poem Generation, Decoding Methods*

## 1. Introduction

Art is a way for humans to express their feelings, memories, and ideas and comes in many forms including paintings, sculptures, novels, poems, etc. For centuries it has been only humans who were capable of creating art but, with recent leaps in the field of generative AI and deep learning, it is possible to generate artistic images, text, music, etc. [1]. Poetry generation using neural networks is an intriguing blend of art and technology. By training a neural network on a vast collection of poetry, it can grasp the patterns and structures of couplet, enabling it to compose its own unique poems. Generative artificial intelligence has emerged as a prominent field of study,

---
\* Corresponding author.

Email Addresses: amirpanahandeh@yahoo.com (Amir Panahandeh)
       asemihanie@gmail.com (Hanie Asemi)
       ac.nourani@azaruniv.ac.ir (Esmail Nourani)

revolutionizing various domains, such as computer vision, natural language processing, and creative arts [2].

Poem generation is one of the challenging tasks because it requires a delicate balance of creativity, emotion, and technical skill. It not only should be grammatically and syntactically correct but also should have meaning, and be fluent and coherent [3]. Despite the huge work done on text generation, a few researches have been conducted on poetry generation and evaluation of the generated poets. Persian poems pose a challenge when it comes to comprehension and interpretation, as they encompass intricate literature and employ various figures of speech like metaphor, simile, paradox, and hyperbole. This complexity can be daunting even for native speakers and professionals [4]–[6]. These special characteristics of Persian poetry make it even more challenging to generate via AI. In this research, a decoder-only transformer model is going to be trained to unconditionally generate rhyming couplets in the style of classical Persian poets.

Poetry generation produces creative content and delivers the it in an aesthetically pleasing manner, usually following a specific structure. Creativity in poetry generation comes from using words in unique ways, employing techniques like alliteration, assonance, and elision. It also involves the use of metaphors, symbolism, and other linguistic devices to convey emotions and sentiments through vivid imagery [7].

Transformer architecture, introduced by Vaswani et al. [8], achieved a significant success in sequence and vision modeling due to its huge improvements in terms of performance and efficiency. When it comes to working effectively with small datasets, a common strategy employed by most transformer models is to utilize a pre-training approach [9]. However, pre-training transformers on large datasets and then fine-tuning them on small datasets can sometimes be burdensome or ineffective [10]. Furthermore, the problems in some domains have little or no connection to the domains of pre-trained models. As a result, pre-training has minimal or no impact on performance in a new domain, especially when the domains are significantly different [11], [12]. Additionally, recent studies showed that small language models can surpass larger models when trained on high quality and specialized data [13], [14].

The idea behind this work is to demonstrate how a compact decoder-only transformer model, along with different decoding methods and applying a novel methodology for choosing temperature in decoding methods, can effectively generate Persian classical poems using minimal data. The main contributions of this work are as follows:

- Utilizing a decoder-only Transformer model to learn and generate unconditioned rhyming couplets in the style of classical Persian poets.
- Proposing a novel decoding method including dynamic temperature inspired by the principles of the simulated annealing optimization algorithm and a simple Anti-LM model.
- Evaluating results with BLEU score and human judges and comparing with an existing large language model. Despite having minimal training data and smaller model size, the experiments have shown promising results in generating high-quality couplets that capture the essence of Persian classical poetry.

The rest of this paper is organized as follows. In section II, we present a brief overview of generative language models, especially in poem generation. Section III describes the outline of the model architecture and the used dataset along with training and decoding methodology. Section IV presents the experiments and results of extensive set of evaluation techniques. Finally, we conclude with a summary of our findings and suggestions for future research in the field of AI-generated poetry in Section V.

## 2. Related Work

In recent times, there has been a growing exploration of the application of Natural Language Processing (NLP) techniques in the area of poetry generation. Notably, new methods of poetry generation have been suggested in different languages, such as employing memory networks [15], [16], Variational AutoEncoder (VAE) [17], reinforcement learning [18], and encoder-decoder transformer mechanism which is adapted by [19] for generating Chinese couplets.

After the revolution of Large Language Models (LLMs), such as GPT and BERT, which have significantly advanced the capabilities of language models in various creative domains, researchers are employing them for poetry generation purpose. For instance, Hämäläinen et al. [20] used a pre-trained language model and fine-tuned them for the task of modern poem generation in French by giving keywords or the first verse. The proposed model follows an encoder-decoder architecture where the encoder is a RoBERTa model and the decoder is a GPT-2 model and their result is evaluated by human judges in typicality, emotionality, and understandability aspects. In a study by Belouadi et al. [21], a new English and German poems dataset is generated and they used a character-level decoder-only model ByGPT5 to generate quatrain poems. Bena et al. [7] proposed a poetry generation model based on the GPT-2 architecture by considering 5 different emotions of anger, anticipation, sadness, joy, and trust to generate creative and emotional poems. Their proposed models have 345M parameters and the average percentage of correctly elicited emotion across four poems in each category was calculated using human reviews. In the research of Nguyen et al. [22], traditional Vietnamese poetry generation has been studied and a new model (SP-GPT2) is proposed to improve the cohesion within the poems. They used the LSTM layer on the top of the GPT2 model to capture the long-term dependency of the output sequence. Translating prose to poem also attracted attention, and a few previous studies have addressed this research area. Khanmohammadi et al. [5] represented a transformer-based model to translating pose to poem in Persian.

In summary, transformer architecture and attention mechanisms have become important methods for generating poetry or couplets. However, their application on small datasets presents significant challenges which will be addressed in this study, and a novel decoding method proposed to improve the quality of the generated couplets.

## 3. Methodology

### 3.1 Model Architecture

This study uses a decoder-only model to generate classical Persian poem couplets. Recently many promising large language models like GPT3, GPT2 and PaLM are employing the decoder-only language model to solve Ses2Seq tasks [23]. By using the decoder-only approach, the language model can be trained on an unlabeled dataset, which is easily obtainable. Also, this architecture allows for a remarkable reduction in the model size and improves performance [23]. In this architecture, the model passes the input text directly to the decoder, which generates the subsequent word tokens in an autoregressive manner and its output is projected by a linear layer to the vocabulary size [24]. The architecture of TPPoet is illustrated in Fig 1.

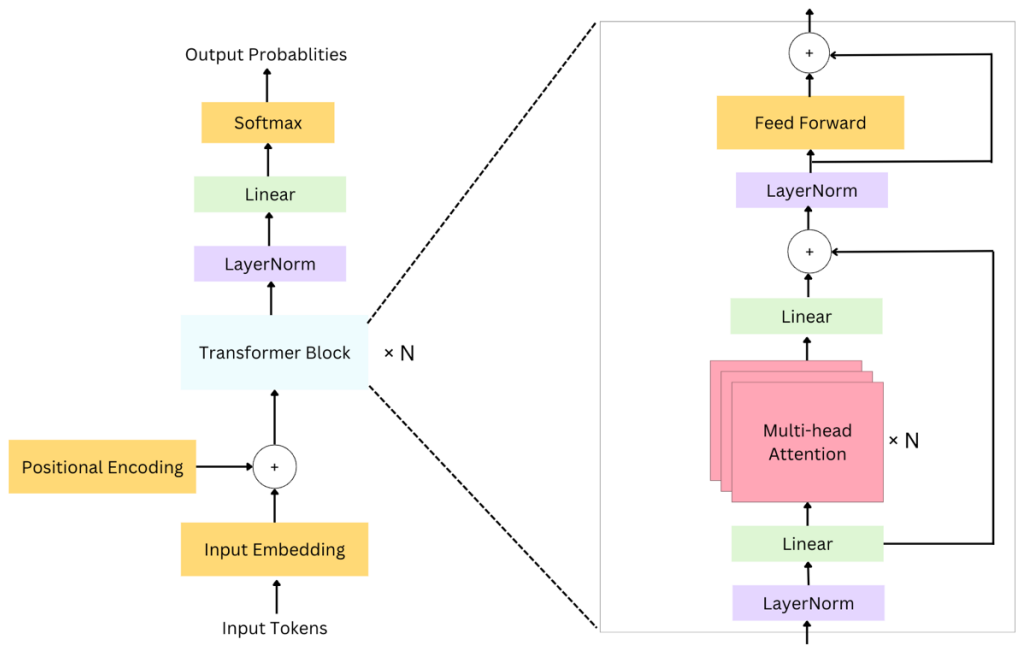

Fig 1. The model architecture of TPPoet

### 3.2 Dataset

For acquiring Persian poem data, we use Ganjoor API[3] to get the list of Persian classic poems. Since these poems have varying lengths with some of them reaching thousands of characters and in order to make training feasible, all poems split into couplets and only rhyming ones are included in the dataset. This results in a dataset of 673743 couplets or ~10 million tokens after tokenizing. After splitting dataset into two training and validation set with 95 to 5 ratio, we are left with 640056 couplets or ~9.5 million tokens for training which is orders of magnitude smaller than data used to pretrain large language models [25].

### 3.3 Tokenization

From bytes to multi-word expressions, text can be analyzed and tokenized in many granularities [26]. In this research, Byte-pair encoding (BPE) is utilized for tokenizing which is a popular technique for building and applying an encoding scheme to natural language texts. BPE traces its origins back to the compression literature, with Gage et al. in [27]. But got popular by [28] in NLP as a tokenizer. BPE is a data-driven approach that begins with an initial base vocabulary and progressively merges the most frequently occurring character pairs in the dataset. These merged pairs are then added to the vocabulary until either the desired vocabulary size is achieved or there are no more pairs left to merge.

---
[3] https://api.ganjoor.net/index.html

In order to choose the most suitable vocabulary size, we conducted experiments with different vocabulary sizes and calculated the average number of tokens in each couplet. As shown in Fig 2 the increase in vocabulary size beyond 8k has minimal impact on the length of our input sequences, but it significantly increases the number of learnable parameters in classification head in TPPoet. The elbow method [29] can be employed to assess the benefits gained from increasing a variable and by utilizing this method, we have selected a vocabulary size of 8k for the remainder of this research.

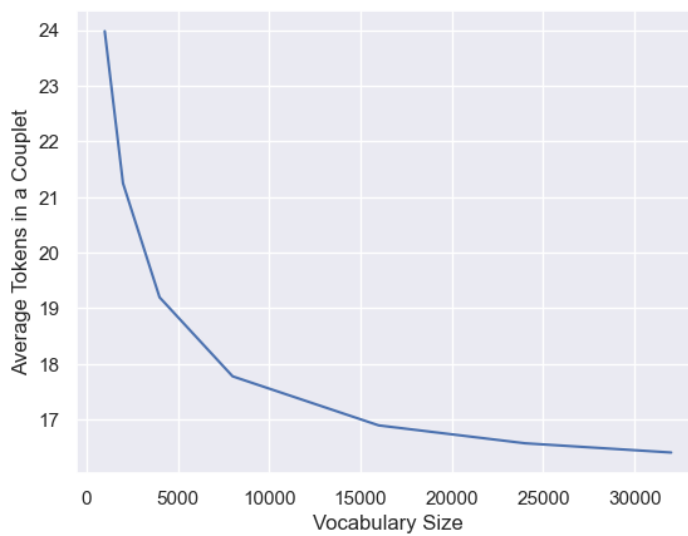

Fig 2. The average tokens in a couplet by changing the vocabulary size. It shows the increase in vocabulary size beyond 10k has a minimal impact on the length of input sequences but causes classification head of decoder to get bigger.

### 3.4 Training

In this study, we conducted our experiment by using different hyper-parameters for the decoder-only vanilla transformer. We included model dimensionality of 384, 512, and 768 which determines vector size representing each token throughout the transformer architecture. Additionally, we varied the number of decoder stacks, considering stacks of 4, 6, and 8. Throughout all our experiments, we ensured consistency by training the models for a maximum of 12 epochs, employing 8 multi-head attention mechanisms, a batch size of 32. In order to capture complex relationships, expansion factor of 4 is used, resulting in 2048 hidden neurons in the first layer of Position-wise Feed-forward block. Table 1 presents the selected hyper-parameters used in our experiment with defined values. These parameters were carefully chosen to explore the impact of different settings on the performance of TPPoet.

Table 1. Learning parameters of the TPPoet

| Hyperparameter | Value |
|---|---|
| Dimension of Model | 384, 512, 768 |
| Number of Decoder Stacks | 4, 6, 8 |
| Epochs | 12 |
| Batch Size | 32 |
| Expansion Factor | 2048 |

During training, we used label smoothing technique [30] with smoothing factor of 0.1 and incorporated Adam [31] as the optimizer with hyperparameters and learning rate scheduling according to [8]:

$$\beta_1 = 0.9, \beta_2 = 0.98, \epsilon = 10^{-9} \tag{1}$$

$$lr = d^{-\frac{1}{2}} \times \min(step^{-\frac{1}{2}}, step \times warmup\_step^{-\frac{3}{2}}) \tag{2}$$

### 3.4 Decoding

The maximization-based decoding strategies that optimize for output with high probability, such as beam search [32] lead to degeneration output text that is bland, incoherent, or gets stuck in repetitive loops even when using big models such as GPT-2 Large [33]. To overcome this issue, many decoding methods proposed like Top-K [34], Nucleus [33]. In this study, we make use of Top-K sampling, limiting the possible words to top 20. In addition to Top-K, we used Nucleus sampling technique with a threshold of 0.9 for expanding and contracting the candidate pool dynamically. Random sampling, by itself, could potentially generate very random words by chance which is not a good way to generate poems as coherence stands as a paramount determinant within the poetic composition. In order to emphasize good candidates and the most probable tokens, a scaling factor called temperature is utilized to shape the distribution before sampling. Therefore, by giving the logits $u_{1:|V|}$ and temperature $t$, the probabilities are re-estimated as:

$$P(x = V_l | x_{1:i-1}) = \frac{\exp(u_l/t)}{\sum_{l'} \exp(\frac{u_{l'}}{t})} \tag{3}$$

In studies, researchers typically choose a temperature value ranging (0,1]. Opting for a lower temperature tends to result in a generation process that prioritizes immediate gains without much room for creativity. Conversely, a higher temperature often leads to incoherent outcomes and, recent analysis has shown that while lowering the temperature improves generation quality, it comes at the cost of decreasing diversity [35]. The establishment of coherence and meaningfulness stands as pivotal factors in the evaluation of high-quality poems, necessitating their careful control. To address this objective within the context of our research, we look at the trade-off between quality and diversity as exploitation and exploration trade-off in optimization problems and propose a novel methodology for choosing temperature inspired by the principles of the simulated annealing optimization algorithm [36]. By having a higher temperature in the beginning, decoder explores the search space can result in higher diversity. As more tokens are added to generated text, temperature decreases and the decoder tempts to choose tokens more relevant both in meaning and suitable to couplet structure. Due to the balance created in exploration and exploitation by using decreasing temperature, it is expected that this technique might result in high quality couplets. Moreover, it is also expected that this change in decoding method, may result in degeneration and repetition due to lower temperature as we move forward in the generation process. In order to overcome this problem, we apply an Anti-LM model [37] to further improve the diversity and quality of generated couplets by penalizing repetitive 2-grams. This simple Anti-LM model can reduce repetition which is a common problem with language models and controlled generation is one of the remedies used in studies [38].

## 4. Results and Discussion

In this section, we analyze the model's hyperparameters and evaluate the proposed decoding strategy described in the previous sections under the selected settings. Finally, we present the results of human evaluation.

### 4.1 Model variants evaluation

In order to choose the right model to be used in the rest of this study, we will test them against validation set and choose the best performing model with small number of learnable parameters. The Fig 3 depicts the validation loss across number of learnable parameters with various hyper-parameter settings. We choose the model with stack of 8 decoders and model dimension of 512 considering that it performs the same as the model with 8 decoders and model dimension of 768 with only 0.3% decrease in validation loss but being 59% of the bigger model size and outperforming the model with 6 decoders and dimensionality of 512 with 0.7% and being only 23% larger.

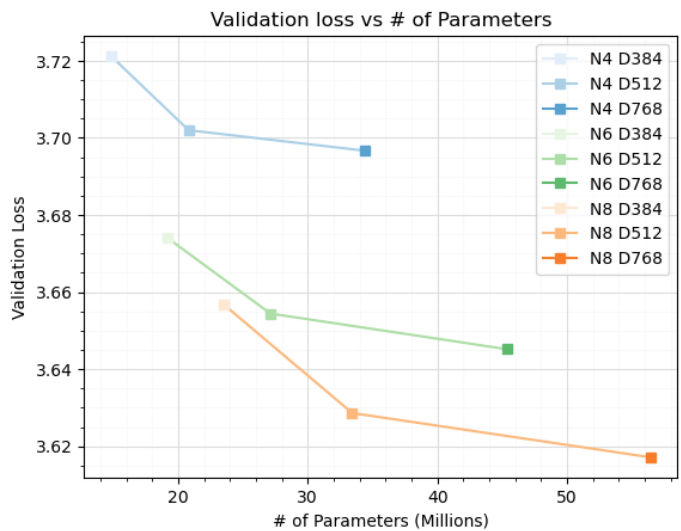

Fig 3. The comparison of the validation loss for the number of learnable parameters resulting from different hyperparameters. In the figure, N represents the number of decoding layers stacked in the decoder model and D represents the model dimension.

## 4.3 Decoding Methods Evaluation

In order to assess the quality of generated couplets, we incorporate BLEU [39], over validation dataset. BLEU score is a widely used metric to evaluate quality of machine-translated text and in the context of unconditional text generation, the test set is considered as reference for generated texts and score is averaged over all references.

Using BLEU score to evaluate generative models and decoding methods is not enough since repetitively generating a single high-quality text over and over can result in a high BLEU score [40]. Self-BLEU [41] is a variation of BLEU metric for diversity evaluation by assessing how each generated text resembles the rest in a generated set. This way, a higher Self-BLEU score shows less diversity and lower values are preferred.

Generating high quality texts without significantly compromising creativity and diversity is a challenging task and is of higher importance especially, in the context of creative text generation. By assessing BLEU and Self-BLEU score over different temperatures while using Nucleus and Top-K sampling, we observe the same effect found by researchers [35] in Fig 4. In this study, by looking at this tradeoff as exploitation and exploration tradeoff in optimization problems, we propose to use simulated annealing technique to decrease temperature by each generated tokens and create a better balance in quality and diversity among generated couplets. Table 2 shows the results of BLEU and Self-BLEU measurements with different decoding techniques and parameters. The two proposed additional techniques to use with Nucleus + Top-K sampling, achieve a high quality without substantially sacrificing diversity and shows the effectiveness of the proposed method.

Table 2. The results of BLUE and Self-BLUE (for 2, 3 and 4 grams) for different decoding techniques and parameters over 1000 generated samples.

| Decoding method | BLEU2 | BLEU3 | BLEU4 | S-BLEU2 | S-BLEU3 | S-BLEU4 |
|---|---|---|---|---|---|---|
| Nucleus + Top-K (t=0.3) | 0.794 | 0.526 | 0.305 | 0.954 | 0.933 | 0.915 |
| Nucleus + Top-K + Anti-LM (t=0.3) | 0.853 | 0.579 | **0.339** | 0.948 | 0.921 | 0.897 |
| Nucleus + Top-K (t=0.5) | 0.841 | 0.580 | 0.33 | 0.838 | 0.715 | 0.611 |
| Nucleus + Top-K + Anti-LM (t=0.5) | 0.864 | 0.6 | 0.336 | 0.839 | 0.712 | 0.6 |
| Nucleus + Top-K (t=0.7) | 0.852 | 0.571 | 0.309 | 0.762 | 0.559 | 0.395 |
| Nucleus + Top-K + Anti-LM (t=0.7) | 0.858 | 0.574 | 0.308 | 0.75 | 0.535 | 0.365 |
| Nucleus + Top-K (t=0.9) | 0.849 | 0.55 | 0.29 | 0.711 | 0.461 | 0.283 |
| Nucleus + Top-K + Anti-LM (t=0.9) | 0.857 | 0.557 | 0.297 | **0.709** | **0.46** | **0.282** |
| Nucleus + Top-K + Anti-LM + Simulated Annealing (T0=0.9, Tf=0.5, step=0.05) (TPPoet) | 0.86 | 0.582 | 0.315 | 0.76 | 0.545 | 0.37 |
| Nucleus + Top-K + Anti-LM + Simulated Annealing (T0=0.9, Tf=0.5, step=0.2) | **0.865** | **0.606** | **0.339** | 0.793 | 0.627 | 0.485 |

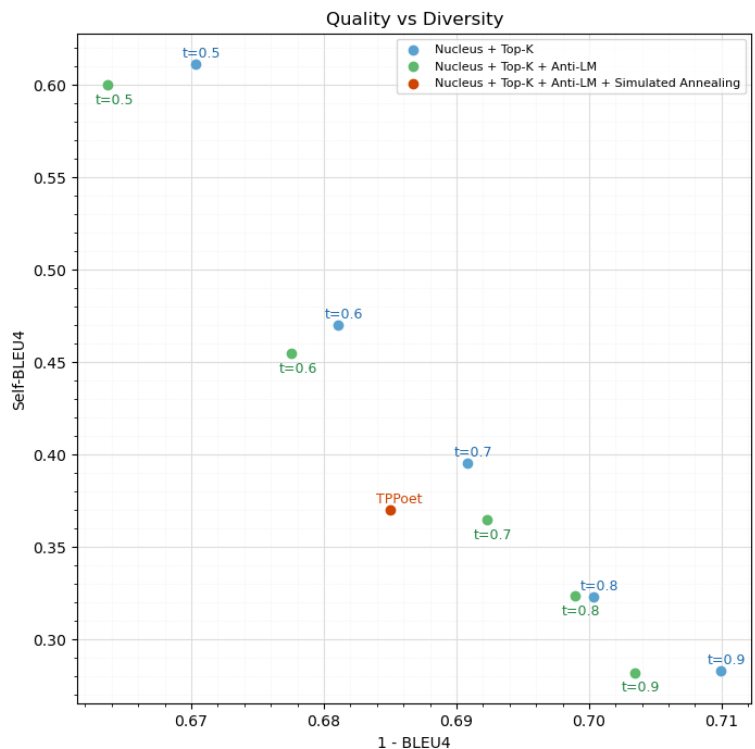

Fig 4. The tradeoff between quality and diversity is shown using 1 - BLEU4 and Self-BLEU4. Temperature sweep from value 0.9 to 0.5 with step of 0.1 for Nucleus + Top-K and Nucleus + Top-K + Anti-LM is drawn along proposed decoding method for different values of temperature decrease rate and, as it can be seen, the proposed Anti-LM model and decreasing temperature in TPPoet decoding method improve diversity and quality.

### 4.4 Human Evaluation

Since automatic evaluation metrics like Perplexity, ROUGE and BLEU can come short in case of creative text generation like poetry [42], [43], in order to evaluate generated couplets and compare them with human poems and GPT2-Persian [44], we conducted a survey containing couplets from our validation set and couplets generated by TPPoet and GPT2-Persian. To handle and eliminate the randomness in the generation process used by both generative models, we repeat each generation 5 times and select the best poem manually. In our survey, we asked participants to assess each couplet with 5 criteria introduced by [3] including fluency, coherence, meaningfulness, poeticness, and Overall from 1 to 5. Additionally, we asked participants to mark each poem with a Human-generated or AI-generated label. A total of 23 participants took part in the evaluation process. For the evaluation, we selected a total of 15 poems to be assessed, with 5 couplets from each generator. The results of the conducted survey are shown in Table 3.

Table 3. The results of human evaluation

| generator | # of Parameters | fluency | coherence | meaningfulness | poeticness | overall | written by human |
|---|---|---|---|---|---|---|---|
| GPT2-Persian [44] | 328M | 2.88 | 2.77 | 2.71 | 2.89 | 2.72 | 30% |
| TPPoet | 33M | **4.34** | **4.12** | **4.08** | **3.96** | **3.94** | **78%** |

| Human | N/A | 3.68 | 3.67 | 3.71 | 3.43 | 3.47 | 66% |

The results demonstrate that our model surpasses the GPT2-Persian model in fluency, coherence, meaningfulness, and poetic qualities. Additionally, our generated poems are identified as human poems 78% on average, a higher rate than the GPT2-Persian model. The presence of meaning and coherence in the poems created by our model has rendered them even more relatable and refined than human poems. Due to the random human couplet selection and non-expert human participants, this may not imply that the TPPoet generated couplets are superior to those of human poets, but rather emphasizes that these generated couplets are not distinguishable from human ones by participants.

## 5. Conclusions

Generating artistic content, despite recent advances, is still a challenging task. In this study, by training a decoder-only transformer with no prior knowledge of Persian language on a relatively small dataset and using an advanced decoding method by combining Nucleus, Top-K, an Anti-LM model and dynamic probability scaling factor (temperature) during generation process, we proposed a novel approach to Persian classical poem generation. Our proposed decoding method showed promising results and automatic evaluations confirmed its ability to generate both diverse and high-quality couplets.

In order to better evaluate the quality of machine generated couplets, we conducted a comparative survey consisting of random couplets from human poets, and produced couplets from GPT2-Persian and TPPoet. Results from human annotators showed that TPPoet was able to generate fluent, meaningful, and human-like poems, outperforming a model ~900% bigger and trained on a huge corpus of general Persian data with capability of generating poems.

In the future, TPPoet can be improved by training on entire poems rather than single rhyming couplets. This approach can enable the model to better grasp the overall meaning and different structures and metres of Persian classical poems. Additionally, sentiment and topic controllable poem generation can be explored in future works.

## 6. References

[1] S. Shahriar, "GAN computers generate arts? A survey on visual arts, music, and literary text generation using generative adversarial network," *Displays*, vol. 73, p. 102237, Jul. 2022, doi: 10.1016/j.displa.2022.102237.

[2] A. Bandi, P. V. S. R. Adapa, and Y. E. V. P. K. Kuchi, "The Power of Generative AI: A Review of Requirements, Models, Input–Output Formats, Evaluation Metrics, and Challenges," *Future Internet*, vol. 15, no. 8, Art. no. 8, Aug. 2023, doi: 10.3390/fi15080260.

[3] H. M. Manurung, "An evolutionary algorithm approach to poetry generation".

[4] A. R. Spofford, "Characteristics of Persian Poetry," *North Am. Rev.*, vol. 140, no. 341, pp. 328–345, 1885.

[5] R. Khanmohammadi, M. S. Mirshafiee, Y. R. Jouryabi, and S. A. Mirroshandel, "Prose2Poem: The Blessing of Transformers in Translating Prose to Persian Poetry." arXiv, Sep. 21, 2022. Accessed: Oct. 09, 2023. [Online]. Available: http://arxiv.org/abs/2109.14934

[6] L. Weitzel, R. C. Prati, and R. F. Aguiar, "The Comprehension of Figurative Language: What Is the Influence of Irony and Sarcasm on NLP Techniques?," in *Sentiment Analysis and Ontology Engineering: An Environment of Computational Intelligence*, W. Pedrycz and S.-M. Chen, Eds., in Studies in Computational Intelligence. , Cham: Springer International Publishing, 2016, pp. 49–74. doi: 10.1007/978-3-319-30319-2_3.

[7] B. Bena and J. Kalita, "Introducing Aspects of Creativity in Automatic Poetry Generation." arXiv, Feb. 06, 2020. doi: 10.48550/arXiv.2002.02511.

[8] A. Vaswani *et al.*, "Attention Is All You Need." arXiv, Aug. 01, 2023. doi: 10.48550/arXiv.1706.03762.

[9] C. Sun, A. Shrivastava, S. Singh, and A. Gupta, "Revisiting Unreasonable Effectiveness of Data in Deep Learning Era," in *2017 IEEE International Conference on Computer Vision (ICCV)*, Oct. 2017, pp. 843–852. doi: 10.1109/ICCV.2017.97.

[10] R. Shao and X.-J. Bi, "Transformers Meet Small Datasets," *IEEE Access*, vol. 10, pp. 118454–118464, 2022, doi: 10.1109/ACCESS.2022.3221138.

[11] S. Mehta, M. Ghazvininejad, S. Iyer, L. Zettlemoyer, and H. Hajishirzi, "DeLighT: Deep and Light-weight Transformer." arXiv, Feb. 11, 2021. doi: 10.48550/arXiv.2008.00623.

[12] Z. Dai, G. Lai, Y. Yang, and Q. V. Le, "Funnel-Transformer: Filtering out Sequential Redundancy for Efficient Language Processing." arXiv, Jun. 05, 2020. doi: 10.48550/arXiv.2006.03236.

[13] S. Gunasekar *et al.*, "Textbooks Are All You Need." arXiv, Oct. 02, 2023. Accessed: Nov. 11, 2023. [Online]. Available: http://arxiv.org/abs/2306.11644

[14] R. Eldan and Y. Li, "TinyStories: How Small Can Language Models Be and Still Speak Coherent English?" arXiv, May 24, 2023. Accessed: Nov. 11, 2023. [Online]. Available: http://arxiv.org/abs/2305.07759

[15] Q. Wang, T. Luo, and D. Wang, "Can Machine Generate Traditional Chinese Poetry? A Feigenbaum Test." arXiv, Jun. 18, 2016. doi: 10.48550/arXiv.1606.05829.

[16] X. Yi, R. Li, and M. Sun, "Generating Chinese Classical Poems with RNN Encoder-Decoder." arXiv, Apr. 06, 2016. doi: 10.48550/arXiv.1604.01537.

[17] X. Yang, X. Lin, S. Suo, and M. Li, "Generating Thematic Chinese Poetry using Conditional Variational Autoencoders with Hybrid Decoders." arXiv, Mar. 05, 2020. doi: 10.48550/arXiv.1711.07632.

[18] X. Yi, M. Sun, R. Li, and W. Li, "Automatic Poetry Generation with Mutual Reinforcement Learning," in *Proceedings of the 2018 Conference on Empirical Methods in Natural Language Processing*, Brussels, Belgium: Association for Computational Linguistics, Oct. 2018, pp. 3143–3153. doi: 10.18653/v1/D18-1353.

[19] Y. Wang, J. Zhang, B. Zhang, and Q. Jin, "Research and Implementation of Chinese Couplet Generation System With Attention-Based Transformer Mechanism," *IEEE Trans. Comput. Soc. Syst.*, vol. 9, no. 4, pp. 1020–1028, Aug. 2022, doi: 10.1109/TCSS.2021.3072153.

[20] M. Hämäläinen, K. Alnajjar, and T. Poibeau, "Modern French Poetry Generation with RoBERTa and GPT-2," arXiv.org. Accessed: Oct. 09, 2023. [Online]. Available: https://arxiv.org/abs/2212.02911v1

[21] J. Belouadi and S. Eger, "ByGPT5: End-to-End Style-conditioned Poetry Generation with Token-free Language Models," arXiv.org. Accessed: Oct. 09, 2023. [Online]. Available: https://arxiv.org/abs/2212.10474v2

[22] T. Nguyen, H. Pham, T. Bui, T. Nguyen, D. Luong, and P. Nguyen, "SP-GPT2: Semantics Improvement in Vietnamese Poetry Generation." arXiv, Oct. 10, 2021. doi: 10.48550/arXiv.2110.15723.

[23] Z. Fu *et al.*, "Decoder-Only or Encoder-Decoder? Interpreting Language Model as a Regularized Encoder-Decoder," arXiv.org. Accessed: Oct. 09, 2023. [Online]. Available: https://arxiv.org/abs/2304.04052v1

[24] M. Fujitake, "DTrOCR: Decoder-only Transformer for Optical Character Recognition," 2023, doi: 10.48550/ARXIV.2308.15996.

[25] H. Naveed *et al.*, "A Comprehensive Overview of Large Language Models." arXiv, Nov. 02, 2023. Accessed: Nov. 11, 2023. [Online]. Available: http://arxiv.org/abs/2307.06435

[26] S. J. Mielke *et al.*, "Between words and characters: A Brief History of Open-Vocabulary Modeling and Tokenization in NLP." arXiv, Dec. 20, 2021. Accessed: Oct. 12, 2023. [Online]. Available: http://arxiv.org/abs/2112.10508

[27] P. Gage, "A new algorithm for data compression," *C Users J. Arch.*, Feb. 1994, Accessed: Oct. 12, 2023. [Online]. Available: https://www.semanticscholar.org/paper/A-new-algorithm-for-data-compression-Gage/1aa9c0045f1fe8c79cce03c7c14ef4b4643a21f8

[28] R. Sennrich, B. Haddow, and A. Birch, "Neural Machine Translation of Rare Words with Subword Units," in *Proceedings of the 54th Annual Meeting of the Association for Computational Linguistics (Volume 1: Long Papers)*, Berlin, Germany: Association for Computational Linguistics, Aug. 2016, pp. 1715–1725. doi: 10.18653/v1/P16-1162.

[29] R. L. Thorndike, "Who belongs in the family?," *Psychometrika*, vol. 18, no. 4, pp. 267–276, Dec. 1953, doi: 10.1007/BF02289263.

[30] C. Szegedy, V. Vanhoucke, S. Ioffe, J. Shlens, and Z. Wojna, "Rethinking the Inception Architecture for Computer Vision." arXiv, Dec. 11, 2015. Accessed: Nov. 11, 2023. [Online]. Available: http://arxiv.org/abs/1512.00567

[31] D. P. Kingma and J. Ba, "Adam: A Method for Stochastic Optimization." arXiv, Jan. 29, 2017. doi: 10.48550/arXiv.1412.6980.

[32] M. Freitag and Y. Al-Onaizan, "Beam Search Strategies for Neural Machine Translation," in *Proceedings of the First Workshop on Neural Machine Translation*, 2017, pp. 56–60. doi: 10.18653/v1/W17-3207.

[33] A. Holtzman, J. Buys, L. Du, M. Forbes, and Y. Choi, "The Curious Case of Neural Text Degeneration." arXiv, Feb. 14, 2020. Accessed: Oct. 12, 2023. [Online]. Available: http://arxiv.org/abs/1904.09751

[34] A. Fan, M. Lewis, and Y. Dauphin, "Hierarchical Neural Story Generation," in *Proceedings of the 56th Annual Meeting of the Association for Computational Linguistics (Volume 1: Long Papers)*, Melbourne, Australia: Association for Computational Linguistics, Jul. 2018, pp. 889–898. doi: 10.18653/v1/P18-1082.

[35] M. Caccia, L. Caccia, W. Fedus, H. Larochelle, J. Pineau, and L. Charlin, "Language GANs Falling Short." arXiv, Feb. 19, 2020. Accessed: Oct. 15, 2023. [Online]. Available: http://arxiv.org/abs/1811.02549

[36] S. Kirkpatrick, C. D. Gelatt, and M. P. Vecchi, "Optimization by Simulated Annealing," *Science*, vol. 220, no. 4598, pp. 671–680, May 1983, doi: 10.1126/science.220.4598.671.

[37] H. Yang, D. Cai, H. Li, W. Bi, W. Lam, and S. Shi, "A Frustratingly Simple Decoding Method for Neural Text Generation." arXiv, May 21, 2023. doi: 10.48550/arXiv.2305.12675.

[38] N. S. Keskar, B. McCann, L. R. Varshney, C. Xiong, and R. Socher, "CTRL: A Conditional Transformer Language Model for Controllable Generation." arXiv, Sep. 20, 2019. doi: 10.48550/arXiv.1909.05858.

[39] B. Chen and C. Cherry, "A Systematic Comparison of Smoothing Techniques for Sentence-Level BLEU," in *Proceedings of the Ninth Workshop on Statistical Machine Translation*, O. Bojar, C. Buck, C. Federmann, B. Haddow, P. Koehn, C. Monz, M. Post, and L. Specia, Eds., Baltimore, Maryland, USA: Association for Computational Linguistics, Jun. 2014, pp. 362–367. doi: 10.3115/v1/W14-3346.

[40] E. Montahaei, D. Alihosseini, and M. S. Baghshah, "Jointly Measuring Diversity and Quality in Text Generation Models." arXiv, May 20, 2019. doi: 10.48550/arXiv.1904.03971.

[41] Y. Zhu *et al.*, "Texygen: A Benchmarking Platform for Text Generation Models." arXiv, Feb. 06, 2018. doi: 10.48550/arXiv.1802.01886.

[42] T. Van de Cruys, "Automatic Poetry Generation from Prosaic Text," in *Proceedings of the 58th Annual Meeting of the Association for Computational Linguistics*, D. Jurafsky, J. Chai, N. Schluter, and J. Tetreault, Eds., Online: Association for Computational Linguistics, Jul. 2020, pp. 2471–2480. doi: 10.18653/v1/2020.acl-main.223.

[43] H. Chen, X. Yi, M. Sun, W. Li, C. Yang, and Z. Guo, "Sentiment-Controllable Chinese Poetry Generation," in *Proceedings of the Twenty-Eighth International Joint Conference on Artificial Intelligence*, Macao, China: International Joint Conferences on Artificial Intelligence Organization, Aug. 2019, pp. 4925–4931. doi: 10.24963/ijcai.2019/684.

[44] bolbolzaban, "gpt2-persian", doi: 10.57967/HF/1207.